\documentclass[runningheads]{llncs}
\usepackage{graphicx}

\usepackage{tikz}
\usepackage{comment}
\usepackage{amsmath,amssymb} 
\usepackage{color}
\usepackage{hyperref}
\usepackage{booktabs}
\usepackage{threeparttable}
\usepackage{makecell}
\usepackage{multirow}
\usepackage{cite}
\usepackage{caption}
\usepackage{subcaption}
\usepackage{ulem} 

\usepackage{xspace} 

\makeatletter
\DeclareRobustCommand\onedot{\futurelet\@let@token\@onedot}
\def\@onedot{\ifx\@let@token.\else.\null\fi\xspace}

\def\ie{\emph{i.e}\onedot}

\makeatother

\usepackage[accsupp]{axessibility}  

\begin{document}
\pagestyle{headings}
\mainmatter

\title{Multi-Granularity Distillation Scheme Towards Lightweight Semi-Supervised Semantic Segmentation} 

\titlerunning{Multi-Granularity Distillation Scheme Towards Lightweight SSSS}
%
\author{Jie Qin\inst{1,2,3}$^{\dagger \star}$\quad   \and
Jie Wu\inst{2}$^{\dagger \ddagger}$\quad   \and
Ming Li\inst{2} \quad \and 
Xuefeng Xiao\inst{2} \and  \\
Min Zheng\inst{2} \and
Xingang Wang\inst{3}$^{\ddagger}$}
\authorrunning{Jie Qin, Jie Wu et al.}
%
\institute{School of Artificial Intelligence, University of Chinese Academy of Sciences \and
ByteDance Inc \and Institute of Automation, Chinese Academy of Sciences}

\renewcommand{\thefootnote}{}
\footnotetext{$^\dagger$Equal contribution. 
$^\star$This work was done while Jie Qin interned at ByteDance. \\
$^\ddagger$Corresponding author. Code is available at \href{https://github.com/JayQine/MGD-SSSS}{github.com/JayQine/MGD-SSSS}.}


\maketitle

\begin{abstract}
Albeit with varying degrees of progress in the field of Semi-Supervised Semantic Segmentation, most of its recent successes are involved in unwieldy models and the lightweight solution is still not yet explored.
We find that existing knowledge distillation techniques pay more attention to pixel-level concepts from labeled data, which fails to take more informative cues within unlabeled data into account.
Consequently, we offer the first attempt to provide lightweight SSSS models via a novel multi-granularity distillation (MGD) scheme, where multi-granularity is captured from three aspects: i) complementary teacher structure; ii) labeled-unlabeled data cooperative distillation;  iii) hierarchical and multi-levels loss setting. 
Specifically, MGD is formulated as a labeled-unlabeled data cooperative distillation scheme, which helps to take full advantage of diverse data characteristics that are essential in the semi-supervised setting.
Image-level semantic-sensitive loss, region-level content-aware loss, and pixel-level consistency loss are set up to enrich hierarchical distillation abstraction via structurally complementary teachers.
Experimental results on PASCAL VOC2012 and Cityscapes reveal that MGD can outperform the competitive approaches by a large margin under diverse partition protocols.
For example, the performance of ResNet-18 and MobileNet-v2 backbone is boosted by \textbf{11.5\%} and \textbf{4.6\%} respectively under 1/16 partition protocol on Cityscapes.
Although the FLOPs of the model backbone is compressed by \textbf{3.4-5.3$\times$} (ResNet-18) and \textbf{38.7-59.6$\times$}  (MobileNetv2), the model manages to achieve satisfactory segmentation results. 
  
\end{abstract}

\section{Introduction}
\label{sec:intro}

\begin{figure}[t!]
  \centering
  \begin{center}
      \includegraphics[width=1.0\linewidth]{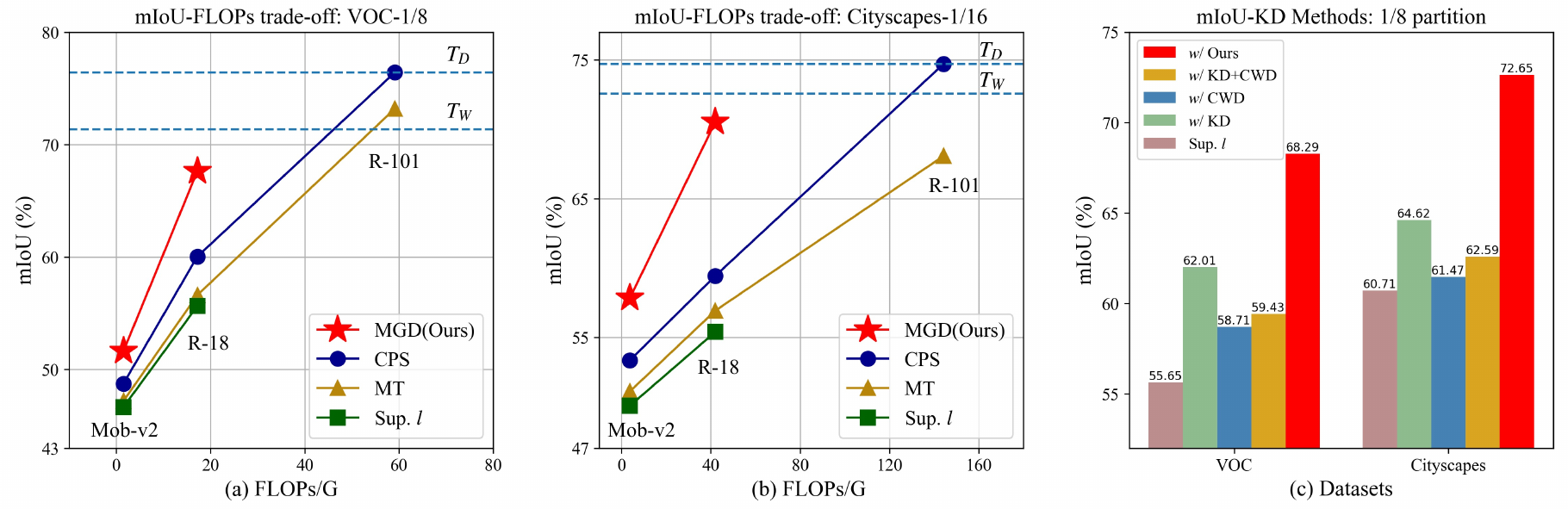}
  \end{center}
    \caption{(a)/(b) The mIoU-FLOPs trade-off between MGD and existing methods including Sup. ${l}$ , MT~\cite{tarvainen2017mean} and CPS~\cite{chen2021semi}. $T_D$ and $T_W$ denote two complementary teacher models. 
  Sup. ${l}$ denotes the model trained only with the labeled data under the corresponding partition. (c) The results of existing distillation methods (KD\cite{wang2020intra} and CWD\cite{shu2021channel}) for optimizing lightweight model of ResNet-18}
  \label{fig: flops}
  
\end{figure}

Semantic segmentation is a fundamental and crucial task due to extensive vision applications such as scene understanding and autonomous driving. However, this task is still extremely dependent on adequate granular pixel-level annotations~\cite{long2015fully, chen2017deeplab, chen2018encoder, zhao2017pyramid}, which requires a huge amount of manual effort.
To alleviate such expensive and unwieldy annotations, some researchers attempt to address this task in the semi-supervised paradigm, where the model is merely accessible to a small amount of image-label pairs joint with abundant unlabeled images.
This is an exceedingly favorable setting since such coarse unlabeled images are more readily available on the internet. The main challenge under this paradigm is how to take good advantage of unlabeled data to alleviate the drastic drop in performance when the labeled data is reduced.
The recent state-of-the-art semi-supervised semantic segmentation approaches~\cite{hung2018adversarial, ouali2020semi, zou2020pseudoseg, chen2021semi} mostly benefit from pseudo labeling and consistency regularization.
%

Despite semi-supervised semantic segmentation (SSSS) having witnessed prevailing success in a series of computer vision applications, most previous approaches~\cite{zou2020pseudoseg, zhong2021pixel, chen2021semi} have resorted to unwieldy and cumbersome models. 
It is arduous to popularize these models that require tremendous computational costs, which becomes a critical bottleneck when such models are required to deploy on resource-constrained server platforms or even more lightweight mobile devices. 
As shown in Fig.~\ref{fig: flops}, the performance of the existing competitive approaches suffers from drastic performance degradation when replacing the cumbersome backbone with a lightweight one such as ResNet-18~\cite{he2016deep} or MobileNet-v2 ~\cite{sandler2018mobilenetv2}. 
We also leverage the existing distillation methods ~\cite{xie2018improving, wang2020intra, shu2021channel} to optimize lightweight models but also get unimpressive results. 
It may due to the fact that existing distillation methods mainly transfer pixel-level knowledge, which lacks diverse and hierarchical distillation signals. Furthermore, these approaches are designed for labeled data and fail to take the data characteristic of unlabeled data into account, hence they are not well adapted to the SSSS task.
These observations motivate us to raise a question:
\textit{how to obtain a lightweight SSSS model with a lighter computation cost joint with high performance?}
Compared with traditional semi-supervised semantic segmentation, lightweight solution confronts three additional challenges:
(i) How to design teacher models to efficiently assist in boosting the capacity of the lightweight model?
(ii) How to fully leverage both limited labeled and sufficient unlabeled data in the semi-supervised distillation setting?
(iii) Are there more granularity concepts that can be resorted to refine the lightweight model for this fine-grained task?

To answer these questions, we propose a novel \textit{\textbf{M}ulti-\textbf{G}ranularity \textbf{D}istillation (MGD) } scheme to obtain a lightweight SSSS model with impressive performance and lite computational costs.  
This scheme elaborately designs triplet-branches to leverage complementary teacher models and hierarchical auxiliary supervision to distill the lightweight model.
Specifically, we capture multi-level concepts to formulate the hierarchical optimization objectives, which can be regarded as diverse granularity supervision signals and play a key role to break through the bottleneck of the capacity of the lightweight student model. The contributions of this work are summarized into four folds:

\begin{itemize}
    \item To the best of our knowledge, we offer the first attempt to obtain the lightweight model for semi-supervised semantic segmentation.
    We provide efficient solutions for the challenges of this newly-raised task and alleviate the issue of performance degradation during model compression.
    We believe our work provides the foundation and direction for lightweight research in the field of semi-supervised semantic segmentation.
  
    \item We propose a novel multi-granularity distillation (MGD) scheme that employs triplet-branches to distill task-specific concepts from two complementary teacher models into a student one. The deep-and-thin and shallow-and-wide teachers help to provide comprehensive and diverse abstractions to boost the lightweight model.
  
      \item The labeled-unlabeled data cooperative distillation is designed to make full use of the characteristics of extensive unlabeled data in the semi-supervised setting. Then we design a hierarchical loss paradigm to provide multi-level ( image-level/region-level/pixel-level) auxiliary supervision for unlabeled data, which supervises the lightweight model to grasp the high-level categorical abstraction and low-level contextual information.  

  \item Extensive experiments on PASCAL VOC2012 and Cityscape demonstrate that MGD can outperform the competitive approaches by a large margin under diverse partition protocols.
Although the FLOPs of the model backbone is compressed by \textbf{3.4-5.3$\times$} (ResNet-18) and \textbf{38.7-59.6$\times$}  (MobileNetv2), the model still can achieve satisfactory performance. 

  \end{itemize}

\section{Related Work}
\label{sec:related}

\subsection{Semantic Segmentation}
Semantic segmentation is a fundamental and crucial task in the field of computer vision~\cite{chen2017deeplab, chen2018encoder, zhao2017pyramid, qin2022activation}. It aims to perform a pixel-level prediction to cluster parts of an image together that belong to the same object class. 
Many approaches~\cite{ronneberger2015u, badrinarayanan2017segnet, noh2015learning} employ the encoder-decoder structure to resume the resolution of features step by step and accomplish pixel-level classification simultaneously. In order to capture the long-range relationships, DeepLab v3+~\cite{chen2017deeplab} make use of dilated convolution~\cite{yu2015multi} to increase the receptive field of networks. ~\cite{yang2018denseaspp, zhao2018icnet, zhao2017pyramid} use the pyramid pooling to capture multi-scale features and integrate pyramid features to obtain global context information. ~\cite{huang2019ccnet, fu2019dual, zhao2018psanet, zhu2019asymmetric, wang2022mvster} use the attention mechanism to aggregate the global contexts that benefit dense prediction. HRNet~\cite{sun2019deep} maintains the context of images by holding the resolution of features. 

\subsection{Semi-Supervised Learning}
Semi-supervised learning aims to utilize limit labeled data and a large amount of unlabeled data to improve the model capacity~\cite{grandvalet2005semi, lee2013pseudo, kipf2016semi, liu2019deep}. The consistency regularization technique applies different perturbations or augmentations and encourages the model to produce invariant outputs~\cite{tarvainen2017mean, xie2019unsupervised, miyato2018virtual}. 
For example, VAT~\cite{miyato2018virtual} applies adversarial perturbations to the output and Temporal Model~\cite{laine2016temporal} incorporates the outputs over epochs and makes them consistent. 
Data augmentations or mixing are performed to generate the different perturbed data to train networks~\cite{berthelot2019mixmatch, berthelot2019remixmatch}. 
FixMatch~\cite{sohn2020fixmatch} takes advantage of the weak augmentation to generate the pseudo labels and supervise the images with strong augmentation.
MT~\cite{tarvainen2017mean} generates the predictions by exponential moving average (EMA). 
Self-supervised learning~\cite{zhai2019s4l},  GAN-based methods~\cite{dai2017good, springenberg2015unsupervised, li2017triple}  and self-training~\cite{xie2020self, dong2018tri} are also exploited to assist semi-supervised learning.

\subsection{Semi-Supervised Semantic Segmentation}
Semi-supervised semantic segmentation is proposed to accomplish the segmentation task with limited image-label pairs and ample unlabeled images.
GAN-based methods~\cite{souly2017semi, hung2018adversarial, mittal2019semi} are proposed to synthesize additional training data or generate the pseudo labels by using adversarial loss on predictions. 
From the perspective of consistency regularization, data augmentations~\cite{zhang2017mixup, yun2019cutmix}, feature space pertubations~\cite{ouali2020semi} and network initialization~\cite{ke2020guided, chen2021semi} are applied to generate the different types of predictions and compute consistency losses.
CCT~\cite{ouali2020semi} exploits the consistent supervision on the outputs of the main decoder and several auxiliary decoders with different perturbations. 
GCT~\cite{ke2020guided} employs two segmentation networks with the same structure but different initialization and enforces the consistency between the predictions.   PC$^{2}$Seg~\cite{zhong2021pixel} and ~\cite{alonso2021semi, lai2021semi} take advantage of the contrastive learning to improve the representation ability on the unlabeled data.
PseudoSeg~\cite{zou2020pseudoseg} uses the weak augmented images to generate the pseudo labels, which are used to supervise the strong augmented images. 

\subsection{Knowledge Distillation}
Knowledge Distillation ~\cite{hinton2015distilling,gou2021knowledge} is a knowledge transfer technique that optimizes a lightweight student model with effective information transfer and supervision of a larger teacher model or ensembles. 
Besides the knowledge transfer in the outputs, the feature maps~\cite{romero2014fitnets, liu2020structured} in the intermediate layers of networks are used to improve the performance of the student networks. Moreover, some methods attempt to transfer the attention concepts~\cite{wu2018image, wu2019pseudo, wu2017global} of feature maps from each channel in intermediate layers. Some approaches try to exploit the knowledge distillation on the dense prediction tasks\cite{xie2018improving, wang2020intra, liu2019structured, shu2021channel}. In addition, ~\cite{you2017learning} indicates that multiple teacher networks can provide more effective information for training a lightweight and ascendant student network. In this paper, we design a deeper and a wider model to play as the complementary teachers and provide multi-granularity auxiliary supervision.

\section{Methodology}
\label{sec: method}
In this section, we first illustrate the overview of the proposed multi-granularity distillation (MGD) framework. Then the complementary teacher distillation and the novel labeled-unlabeled data cooperative distillation are presented in Sec.~\ref{sec: comp_tea} and Sec.~\ref{sec: label}, respectively. The hierarchical and multi-level distillation for unlabeled data is introduced Sec.~\ref{sec: multi_task}.  

\begin{figure*}[t!]
  \centering
  \begin{center}
      \includegraphics[width=0.95\textwidth]{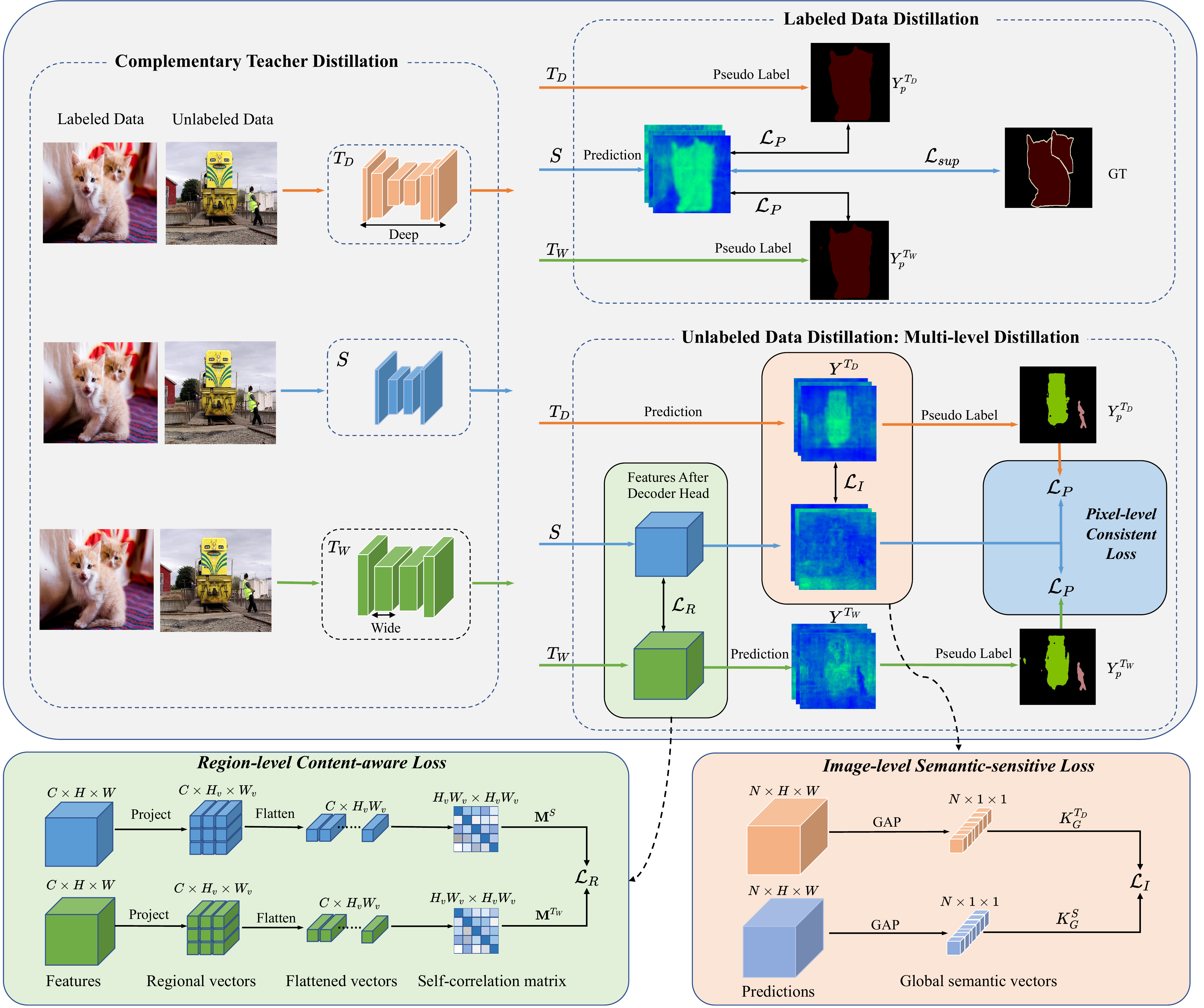}
  \end{center}
  \caption{The pipeline of Multi-Granularity Distillation (MGD) Scheme. MGD is formulated as triplet-branches, where there are two structurally-complementary teacher models (\ie., $T_D$ and $T_W$) and a target student model $S$. $T_D$ indicates the deep-and-thin teacher model, $T_W$ denotes the shallow-and-wide teacher model, and the student network $S$ is a shallow and thin model. Image-level semantic-sensitive loss, region-level content-aware loss, and pixel-level consistency loss are designed to form the hierarchical and multi-level loss setting}
  
  \label{fig: framework}
\end{figure*}

\subsection{Multi-granularity Distillation Scheme}
\label{sec: scheme}
Due to limited labeled data and restricted learning capacity of lightweight models, conventional consistency regularization approaches ~\cite{zou2020pseudoseg, ouali2020semi, chen2021semi} suffer from a drastic degradation in performance when they are directly applied on the lightweight models (refer to Fig.~\ref{fig: flops}).
Furthermore, we also tried a series of existing distillation methods to relieve significant performance degradation but found that they are not so effective under the setting of semi-supervised segmentation.
Therefore, we propose a multi-granularity distillation scheme to leverage task-oriented knowledge distillation techniques to guide the learning procedure of the lightweight model, as shown in Fig.~\ref{fig: framework}. 
Specifically, multi granularity denotes that the distillation process is carried out synergistically from diverse aspects. Firstly, we design dual teachers to provide complementary distillation abstractions. Secondly,  we formulate a labeled-unlabeled data cooperative distillation to optimize the network via the data characteristics of semi-supervision. Thirdly, we employ image-level, region-level and pixel-level visual cues from complementary teacher models to formulate a hierarchical and multi-level optimization objective.

\subsection{Complementary Teacher Distillation}
\label{sec: comp_tea}
\noindent \textit{Solution for ``How to design teacher models to efficiently assist in boosting the capacity of the lightweight model?"}

Although \cite{you2017learning} has proved that multiple teacher models can provide more effective information for students, the research on diverse teacher structures is still insufficient.
For the fine-grained task of semantic segmentation, complementary teacher networks can bring two advantages: i) help to capture more comprehensive visual cues and boost the performance of the student model from different aspects; ii) suppress over-fitting issue that is common in the distillation. Hence in this paper, the teacher model is expanded from two complementary dimensions compared to the student model, \ie, depth and width. Specifically, the complementary teachers contain a deep-and-thin teacher model  $T_D$  (such as ResNet-101~\cite{he2016deep}) and a wide-and-shallow teacher model $T_W$ (such as Wide ResNet-34~\cite{zagoruyko2016wide}).
These two complementary teacher models can achieve comparable performance while tending to contain diverse potential characteristics.

$T_D$ has proven its ability in extracting high-level semantic and global categorical abstractions\cite{szegedy2015going}, which helps to achieve promising results in the classification-oriented task. Hence, we attempt to extract the global semantic knowledge $K_G^{T_D}$ from $T_D$ to enhance the global perception ability of the student model. 
Furthermore, $T_W$ is conducive to capturing the diverse content-aware information due to the sufficient channels, which is beneficial for modeling the contextual relationship between regions.
In this paper, we first follow CPS \cite{chen2021semi} to pre-train two teacher models and freeze their parameters to conduct an offline distillation procedure for refining the student model.

\subsection{Labeled-unlabeled Data Cooperative Distillation}
\label{sec: label}
\noindent \textit{Solution for ``How to fully leverage both limited labeled and sufficient unlabeled data in the semi-supervised distillation setting?"}

The existing distillation techniques are mainly carried out under the full supervision setting and lack of distillation practice for unlabeled data. 
So we involve a novel labeled-unlabeled data cooperative distillation scheme in MGD.
As for labeled data ($X_l$) distillation, the student model is optimized by ground truth labels $\hat{Y}$ with the supervision loss $\mathcal{L}_{Sup}^{l}$:
\begin{equation}
    \mathcal{L}_{Sup}^{l}(Y, \hat{Y}) = -\frac{1}{H \times W} \sum_{i=1}^{H \times W}\left(\hat{y_i} log(y_{i}))\right. \text{,}
    \end{equation}
where $y_i$ represents the prediction of the student model and the $\hat{y_{i}}$ illustrates the ground truth label. Superscript $l$ represents labeled data.

Furthermore, we design the pixel-level consistency loss to maintain that the multiple predictions of the same input are consistent. 
Specifically, both output of the teacher model $T_D$ and $T_W$ can be regarded as pseudo label $Y_p$ for the student model $S$. Pixel-level consistency loss is set as:
\begin{equation}
\begin{aligned}
  \mathcal{L}_{P}^{l}(Y, Y_p) &= \mathcal{L}_{ce}(Y, Y_p^{T_D}) + \mathcal{L}_{ce}(Y, Y_p^{T_W}) \\
  &= -\frac{1}{H \times W} \sum_{i=1}^{H \times W}[y_{pi}^{T_D} log(y_i)+ y_{pi}^{T_W} log(y_i)]\text{,}
  \end{aligned}
  \end{equation}
where $y_{pi}^{T_D}$ and $y_{pi}^{T_W}$ are the pseudo labels generated by $T_D$ and $T_W$ respectively. $H \times W$ denotes the number pixels of prediction masks. 
The $\mathcal{L}_{Sup}^{l}$ and $\mathcal{L}_{P}^{l}$ are combined to form the labeled distillation loss $\mathcal{L}_{label}$. 
Besides labeled data distillation, we further design a hierarchical and multi-level loss setting for unlabeled data distillation $\mathcal{L}_{unlabel}$ in Sec.~\ref{sec: multi_task}.

\subsection{Hierarchical and Multi-level Distillation}
\label{sec: multi_task}
\noindent \textit{Solution for ``Are there more granularity concepts that can be resorted to refine the lightweight model for this fine-grained task?"}

To capture the multi-granularity features oriented for segmentation task, we design a hierarchical and multi-level distillation scheme for unlabeled data, which helps to provide students with holistic and local concepts in various aspects.
Because there is no ground-truth supervision for unlabeled data, we introduce image-level semantic-sensitive loss, region-level content-aware loss, and pixel-level consistency loss to achieve a more comprehensive distillation procedure on unlabeled data.

\textbf{Image-level Semantic-sensitive Loss.}
Image-level semantic-sensitive loss is introduced to distill the high-dimensional semantic abstraction from the deeper teacher model $T_D$ into the lightweight one. After obtaining the predictions $Y \in \mathcal{R}^{N \times H \times  W}$ from $T_D$, we employ global average pooling (GAP) operation to calculate the global semantic vector of each category, which is illustrated as:
\begin{equation}
  K_G^{T_D}=G(Y^{N \times H \times  W}) \text{,}
  \end{equation}
where $K_G^{T_D} \in \mathcal{R}^{N \times 1 \times  1}$ summarizes the global semantic knowledge of $N^{th}$ categories. $G$ denotes the global average pooling operation in each channel. We further calculate the global semantic representation $K_G^{S}$ of student model and employ image-level semantic-sensitive loss $\mathcal{L}_{I}^{u}$ to supervise it:
\begin{equation}
  \mathcal{L}_{I}^{u}\left(K_{G}^{S}, K_{G}^{T_{D}}\right)=\frac{1}{N}\sum_{i=1}^{N}\Vert {k_{Gi}^{S}-k_{Gi}^{T_{D}}}\Vert_1 \text{,}
  \end{equation} 
where $k_{Gi}^{S}$ and $k_{Gi}^{T_{D}}$ denote the semantic category of the student $S$ and the teacher $T_D$ respectively. $N$ represents the number of categories and $u$ represents unlabeled data.
In this way, the student model attempts to learn higher-dimensional semantic representations, which helps to provide global guidance on distinguishing task-specific categories with respect to the semantic segmentation task.


\textbf{Region-level Content-aware Loss.}
Region-level content-aware loss aims to take advantage of channels advantage of wider teacher model $T_W$ to provide plentiful regional context information. It leverages correlation between image patches from $T_W$ to guide the student $S$ to model contextual relationships between regions, which is effective to better understand image regional details and relationships. The decoders output features  ($F^{T_W}$ and $F^S$)  from $T_W$ and $S$ are extracted to calculate the  content-aware loss. Firstly, $F \in \mathcal{R}^{C \times H \times  W}$ are projected into the regional content vectors $\mathbf{V} \in \mathcal{R}^{C \times H_v \times  W_v}$ via projection heads. Each vector $\boldsymbol{v} \in \mathcal{R}^{C \times 1 \times  1}$ in $\mathbf{V}$ represents the regional content of the patch features with size $C \times H/H_v \times  W/W_v$. The projection heads are employed with the adaptive average pooling. Then, the regional content vectors $\mathbf{V}$ are used to calculate the self-correlation matrix $\mathbf{M} \in \mathcal{R}^{ H_vW_v \times H_vW_v}$:
\begin{equation}
  \mathbf{M}=sim(\mathbf{V}^T\mathbf{V}) \text{,} \ with \ 
  {m}_{ij}=\frac{\boldsymbol{v}_i^T \boldsymbol{v}_j}{\Vert  \boldsymbol{v}_i \Vert  \Vert  \boldsymbol{v}_j \Vert } \text{,}
  \end{equation} 
where ${m}_{ij}$ denotes the value of self-correlation matrix at the coordinate of row $i$ and column $j$ produced with the cosine similarity $sim()$. The $\boldsymbol{v}_i$ and $\boldsymbol{v}_j$ are the $i^{th}$ and $j^{th}$ vectors of flattened $\mathbf{V} \in \mathcal{R}^{C \times H_vW_v}$. 
The self-correlation matrix represents the regional relationship among features $F$, which can reflect the correlation between image regions.
The region-level content-aware loss is set up to minimize the difference between $\mathbf{M}^{T_{W}}$ and $\mathbf{M}^S$: 
\begin{equation}
  \mathcal{L}_{R}^{u}\left(\mathbf{M}^S, \mathbf{M}^{T_{W}} \right)=\frac{1}{ H_vW_v \times H_vW_v}\sum_{i=1}^{H_vW_v}\sum_{j=1}^{H_vW_v}({m_{ij}^{S}-{m}_{ij}^{T_{W}}})^2  \text{,}
  \end{equation} 
where $m_{ij}^{S}$ and $m_{ij}^{T_{W}}$ denote the self-correlation values of student model and teacher model. $H_vW_v \times H_vW_v$ is the size of self-correlation matrices.

\textbf{Pixel-level Consistency Loss.}
The pixel-level consistency loss is also employed on the unlabeled data $X_u$, which is the same as Eq.(2) and denoted as $\mathcal{L}_{P}^{u}$. To sum up, the total loss function in our multi-granularity distillation scheme is summarized as: 
\begin{equation}
    \mathcal{L} = \underbrace{\mathcal{L}_{Sup}^{l} + \mathcal{L}_{P}^{l}}_{\mathcal{L}_{label}} + \underbrace{\mathcal{L}_{P}^{u} + \lambda_1 \mathcal{L}_{I}^{u} + \lambda_2 \mathcal{L}_{R}^{u}}_{\mathcal{L}_{unlabel}} \text{,}
    \end{equation}
where the $\lambda_1$ and $\lambda_2$ are the trade-off parameters for the loss functions.

\section{Experiment}
\label{sec: exp}

\subsection{Datasets and Evaluation Metrics}
\textbf{Datasets.} PASCAL VOC2012~\cite{everingham2010pascal} is a widely used benchmark for semantic segmentation, which consists of 20 foreground objects classes and one background class. Following ~\cite{ke2020guided, chen2021semi}, we adopt 10,582 augmented images as the full training set and 1,449 images for validation. Cityscapes~\cite{cordts2016cityscapes} is a real urban scene dataset that consists of 2,974 images for training, 500 images for validation, and 1,525 for testing with 19 semantic classes. We follow ~\cite{ke2020guided, chen2021semi} to use the public 1/16, 1/8, 1/4, and 1/2 subsets of the training set as the labeled data and the remaining images in the training set as the unlabeled data.

\textbf{Evaluation Metrics.} 
To verify the performance of the segmentation model, we calculate the mean Intersection-over-Union (mIoU) of all classes. We also compare the parameters and FLOPs of the model backbone for fair evaluation.

\subsection{Implementation Details}
We initialize the weights of the lightweight backbones (ResNet-18~\cite{he2016deep} and MobileNetv2~\cite{sandler2018mobilenetv2}) with pre-trained weights on ImageNet~\cite{krizhevsky2012imagenet}. We employ DeepLabv3+ \cite{chen2018encoder} segmentation architecture in our framework, which consists of a visual backbone and an ASPP based segmentation heads. The deep-and-thin teacher model ResNet-101 and the wide-and-shallow teacher model Wide ResNet-34\cite{zagoruyko2016wide}  are pre-trained 
via CPS~\cite{chen2021semi} and do not update their parameters in the procedure of optimizing student models. We leverage SGD algorithm for network optimization with a 0.0005 weight decay and a 0.9 momentum. We set the hyper-parameters of the coefficients of the loss functions as $\lambda_1 = 0.002$ and $\lambda_2 = 100$. There is currently no researches on lightweight semi-supervised semantic segmentation. 
We replace the backbone of competitive methods \ie, CPS~\cite{chen2021semi} and MT~\cite{tarvainen2017mean} with ResNet-18 and MobileNetv2 to make a comparison with our approach. The classical distillation methods KD~\cite{wang2020intra} (KL Divergence) and CWD~\cite{shu2021channel} are employed on our complementary teacher models to distill the lightweight models.

\begin{table*}[t!]
  \centering
  \caption{Comparison with SOTAs on the VOC2012 Val set with $512 \times 512$ input size under different partition protocols. The first split reports the results of the complementary teacher models. The second and third splits denote the results of employing ResNet-18 and MobileNetv2 as the lightweight backbone, respectively.
  Sup. ${l}$ denotes training with only labeled data under the corresponding partition.}
  \begin{threeparttable}
  \resizebox{\linewidth}{!}{
      \begin{tabular}{lcccccccc}
      \toprule
      \textbf{Method} & \textbf{Backbone} & \textbf{Params} & \textbf{FLOPs}  & \textbf{1/16 (662)} & \textbf{1/8 (1323)} & \textbf{1/4 (2646)} & \textbf{1/2 (5291)} \\ 
      \midrule
      CPS~\cite{chen2021semi} & ResNet-101 ($T_D$) & 42.63M & 59.13G & 73.98 & 76.43 & 77.61 & 78.64   \\
      CPS~\cite{chen2021semi} & Wide ResNet-34 ($T_W$) & 66.83M & 91.12G & 68.17 & 71.38 & 74.78 & 75.85   \\
      \midrule
      Sup. ${l}$ & ResNet-18 & \multirow{4}{*}{11.19M} & \multirow{4}{*}{17.20G} & 42.17 & 55.65 & 62.52 & 65.39   \\
      MT~\cite{tarvainen2017mean} & ResNet-18 & \multirow{4}{*}{\footnotesize{(\textbf{3.81$\times$} w.r.t. $T_D$)}} & \multirow{4}{*}{\footnotesize{(\textbf{3.44$\times$} w.r.t. $T_D$)}} & 44.61 & 56.62 & 61.79 & 66.48   \\
      CPS~\cite{chen2021semi} & ResNet-18 & \multirow{4}{*}{\footnotesize{(\textbf{5.97$\times$} w.r.t. $T_W$)}} & \multirow{4}{*}{\footnotesize{(\textbf{5.30$\times$} w.r.t. $T_W$)}} & 47.13 & 60.05 & 65.89 & 68.69   \\
      KD~\cite{wang2020intra} & ResNet-18 & & & 56.37 & 62.01 & 66.23 & 68.55 & \\
      CWD~\cite{shu2021channel} & ResNet-18 & & & 50.45 & 58.71 & 63.96 & 65.82 & \\
      Ours (MGD) & ResNet-18 & &  & \textbf{66.86} & \textbf{68.29} & \textbf{69.71} & \textbf{71.59}   \\
      \midrule
      Sup. ${l}$ & MobileNetv2 & \multirow{4}{*}{1.81M} & \multirow{4}{*}{1.52G} & 46.02 & 46.66 & 48.19 & 49.61   \\
      MT~\cite{tarvainen2017mean} & MobileNetv2 & \multirow{4}{*}{\footnotesize{(\textbf{23.55$\times$} w.r.t. $T_D$)}} & \multirow{4}{*}{\footnotesize{(\textbf{38.90$\times$} w.r.t. $T_D$)}} & 45.11 & 47.20 & 49.15 & 49.63  \\
      CPS~\cite{chen2021semi} & MobileNetv2 & \multirow{4}{*}{\footnotesize{(\textbf{36.92$\times$} w.r.t. $T_W$)}} & \multirow{4}{*}{\footnotesize{(\textbf{59.95$\times$} w.r.t. $T_W$)}} & 46.57 & 48.75 & 49.94 & 50.52  \\
      KD~\cite{wang2020intra} & MobileNetv2 & & & 47.99 & 49.30 & 50.07 & 50.68 \\
      CWD~\cite{shu2021channel} & MobileNetv2 & & & 46.13 & 47.45 & 48.89 &  49.70 \\
      Ours (MGD) & MobileNetv2 &  &  & \textbf{50.95} & \textbf{52.27} & \textbf{53.92} & \textbf{55.88}   \\
      \bottomrule
      \end{tabular}}
      \end{threeparttable}
      \label{tab: res_1}
      
  \end{table*}  
  
\subsection{Comparison with state-of-the-art methods}
\begin{figure}[t]
  \centering
  \begin{center}
      \includegraphics[width=0.9\linewidth]{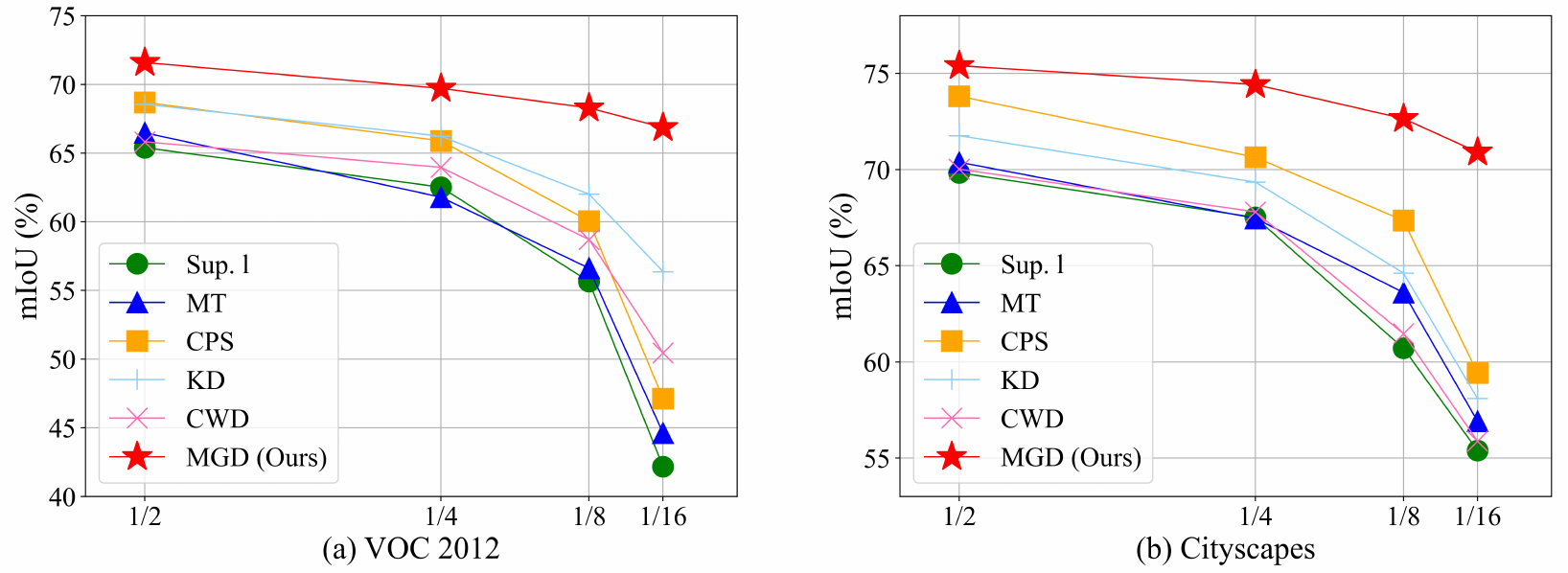}
  \end{center}
  
  \caption{Performance of different methods under different partition protocols with ResNet-18 backbone}
  \label{fig: semi-seg}
  
\end{figure}

In this section, we compare MGD with several state-of-the-art methods in terms of computation cost and accuracy under all partition protocols. We employ ResNet-18 and MobileNetv2 as the lightweight backbones and report the results on PASCAL VOC2012 and Cityscapes in Tab.~\ref{tab: res_1},~\ref{tab:few},~\ref{tab: res_2} and Fig.~\ref{fig: semi-seg}. 
We summarize the FLOPs and parameters of the model backbone, and the corresponding segmentation heads in these models are compressed proportionately.

\textbf{PASCAL VOC2012.}
The experimental results on PASCAL VOC2012 are shown in Tab.~\ref{tab: res_1}, which can be summarized as the following observations:
1) Replacing the large model with a small model directly will result in obvious performance degradation. For example, it decreased from 73.97\% (ResNet-101) to 47.13\% (ResNet-18) under 1/16 partition protocols, with a serious decline of 26.8\%.
2) Employing the existing distillation method obtain minor improvement, while our approach contributes to alleviating the performance drop effectively. 
As shown in Fig.~\ref{fig: semi-seg}, MGD shows significant performance improvement under diverse partition protocols, and the growth trend is more obvious when the amount of labeled data is less.
Compared with CPS, MGD helps to boost the performance from 47.13 to 66.86 in the 1/16 partition setting. In more detail, the improvement is 19.73\%, 8.24\%, 3.82\%, and 2.90\% under 1/16, 1/8, 1/4, and 1/2 partition protocols, respectively.
3) Furthermore, although the model with ResNet-18 backbone is compressed by \textbf{3.44$\times$ (5.30$\times$)} FLOPs and \textbf{3.81$\times$ (5.97$\times$)} parameters, it only exists a small gap from the result of $T_W$. It fully demonstrates the effectiveness of our method.
4) Although the lightweight model with MobileNetv2 compresses more calculational costs (\textbf{38.90$\times$ (59.95$\times$)} FLOPs and \textbf{23.55$\times$ (36.92$\times$)} parameters), MGD manages to achieve an impressive performance. For example, MGD with MobileNetv2 (50.95) even surpasses the CPS method with ResNet-18 backbone (47.13) under 1/16 partition protocol.

\begin{table}[tp]
  \centering
  \footnotesize
  \caption{Comparison for few-supervision on PASCAL VOC2012. We follow the same partition protocols as PseudoSeg ~\cite{zou2020pseudoseg}.}
  \begin{threeparttable}
  \resizebox{0.7\textwidth}{!}{
    \begin{tabular}{lccccc}
      \toprule
      \textbf{Method} & \textbf{Backbone} & \textbf{1/16 (92)} & \textbf{1/8 (183)} & \textbf{1/4 (366)} & \textbf{1/2 (732)} \\ 
      \midrule
      CPS~\cite{chen2021semi} & ResNet-101 & 64.07 & 67.42 & 71.71 & 75.88 \\
      CPS~\cite{chen2021semi} & Wide ResNet-34 & 60.01 & 63.32 & 67.84 & 72.03 \\
      \midrule
      Sup. ${l}$ & ResNet-18 & 15.54 & 22.43 & 35.78 & 51.47 \\
      MT~\cite{tarvainen2017mean} & ResNet-18 & 12.58 & 18.15 & 33.71 & 50.02 \\
      CPS~\cite{chen2021semi} & ResNet-18 & 24.53 & 36.19 & 47.61 & 56.85 \\
      Ours  & ResNet-18 & \textbf{58.03} & \textbf{61.94}  & \textbf{64.97} & \textbf{67.40} \\
      \bottomrule
      \end{tabular}}
      \end{threeparttable}
      \label{tab:few}
  \end{table}

\begin{table*}[t]
    \centering
    \caption{Comparison with state-of-the-arts on the Cityscapes val set with the input size of $800 \times 800$ under different partition protocols.}
    \begin{threeparttable}
    \resizebox{\linewidth}{!}{
        \begin{tabular}{lcccccccc}
        \toprule
        \textbf{Method} & \textbf{Backbone} & \textbf{Params} & \textbf{FLOPs} & \textbf{1/16 (662)} & \textbf{1/8 (1323)} & \textbf{1/4 (2646)} & \textbf{1/2 (5291)}  \\ 
        \midrule
        CPS~\cite{chen2021semi} & ResNet-101($T_D$) & 42.63M & 144.32G & 74.72 & 77.62 & 78.85 & 79.58  \\
        CPS~\cite{chen2021semi} & Wide ResNet-34($T_W$) & 66.83M & 222.45G & 72.56 & 75.41 & 77.68 & 78.37   \\
        \midrule
        Sup. ${l}$ & ResNet-18 & \multirow{3}{*}{11.19M} & \multirow{3}{*}{42.01G} & 55.39 & 60.71 & 67.51 & 69.83  \\
        MT~\cite{tarvainen2017mean} & ResNet-18 & \multirow{3}{*}{\footnotesize{(\textbf{3.81$\times$} w.r.t. $T_D$)}} & \multirow{3}{*}{\footnotesize{(\textbf{3.43$\times$} w.r.t. $T_D$)}} & 56.92 & 63.61 & 67.46 & 70.38  \\
        CPS~\cite{chen2021semi} & ResNet-18 & \multirow{3}{*}{\footnotesize{(\textbf{5.97$\times$} w.r.t. $T_W$)}} & \multirow{3}{*}{\footnotesize{(\textbf{5.30$\times$} w.r.t. $T_W$)}} & 59.43 & 67.35 & 70.64 & 73.81  \\
        KD~\cite{wang2020intra} & ResNet-18 & & & 58.10 & 64.62 & 69.35 & 71.76 \\
        CWD~\cite{shu2021channel} & ResNet-18 & & & 55.88 & 61.47 & 67.80 & 70.02 \\
        Ours (MGD) & ResNet-18 &  &  & \textbf{70.91} & \textbf{72.65} & \textbf{74.42} & \textbf{75.40}  \\
        \midrule
        Sup. ${l}$ & MobileNetv2 & \multirow{3}{*}{1.81M} & \multirow{3}{*}{3.73G} & 50.06 & 51.97 & 54.03 & 56.86  \\
        MT~\cite{tarvainen2017mean} & MobileNetv2 & \multirow{3}{*}{\footnotesize{(\textbf{23.55$\times$} w.r.t. $T_D$)}} & \multirow{3}{*}{\footnotesize{(\textbf{38.69$\times$} w.r.t. $T_D$)}} & 51.10 & 53.85 & 56.21 & 57.25  \\
        CPS~\cite{chen2021semi} & MobileNetv2 & \multirow{3}{*}{\footnotesize{(\textbf{36.92$\times$} w.r.t. $T_W$)}} & \multirow{3}{*}{\footnotesize{(\textbf{59.64$\times$} w.r.t. $T_W$)}} & 53.34 & 55.61 & 57.98 & 59.20  \\
        KD~\cite{wang2020intra} & ResNet-18 & & & 53.45 & 54.89 & 55.28 & 57.90 \\
        CWD~\cite{shu2021channel} & ResNet-18 & & & 51.52 & 52.64 & 54.37 & 57.01 \\
        Ours (MGD)& MobileNetv2 &  &  & \textbf{58.00} & \textbf{59.27} & \textbf{61.36} & \textbf{62.55}  \\
        \bottomrule
        \end{tabular}}
        \end{threeparttable}
        \label{tab: res_2}
    \end{table*}

\textbf{Few Supervision on PASCAL VOC2012.} 
To evaluate the robustness of our method on the few labeled case, we conduct the experiments on PASCAL VOC 2012 with few supervision by following the same partition protocols adopted in PseudoSeg~\cite{zou2020pseudoseg}. 
As shown in Tab.~\ref{tab:few}, it can observe that our method still can achieve impressive results with fewer labeled samples, which far exceed MT~\cite{tarvainen2017mean} and CPS~\cite{chen2021semi}. 
Surprisingly, the performance of MGD (58.03) trained with only 1/16 partition data can surpass CPS (56.85) under 1/2 partition protocols.
The few supervision experiments further reveal that MGD is suitable for lightweight model optimization with few or even very few samples.

\textbf{Cityscapes.}
We conduct the experiments under different partition protocols on Cityscapes and summarize the result in Tab.~\ref{tab: res_2}. 
From the Tab.~\ref{tab: res_2} and Fig.~\ref{fig: semi-seg}, we observe a similar performance gain and trend under different partition protocols as PASCAL VOC2012.
Our method can establish state-of-the-art results among all partition protocols with both ResNet-18 and MobileNetv2 backbones. 
For example, MGD obtains 70.91/58.00 under the 1/16 partition protocol with ResNet-18/MobileNetv2 backbone, which outperforms CPS by 11.48\%/4.66\%.

\begin{figure*}[t!]
\centering
\begin{center}
    \includegraphics[width=0.9\textwidth]{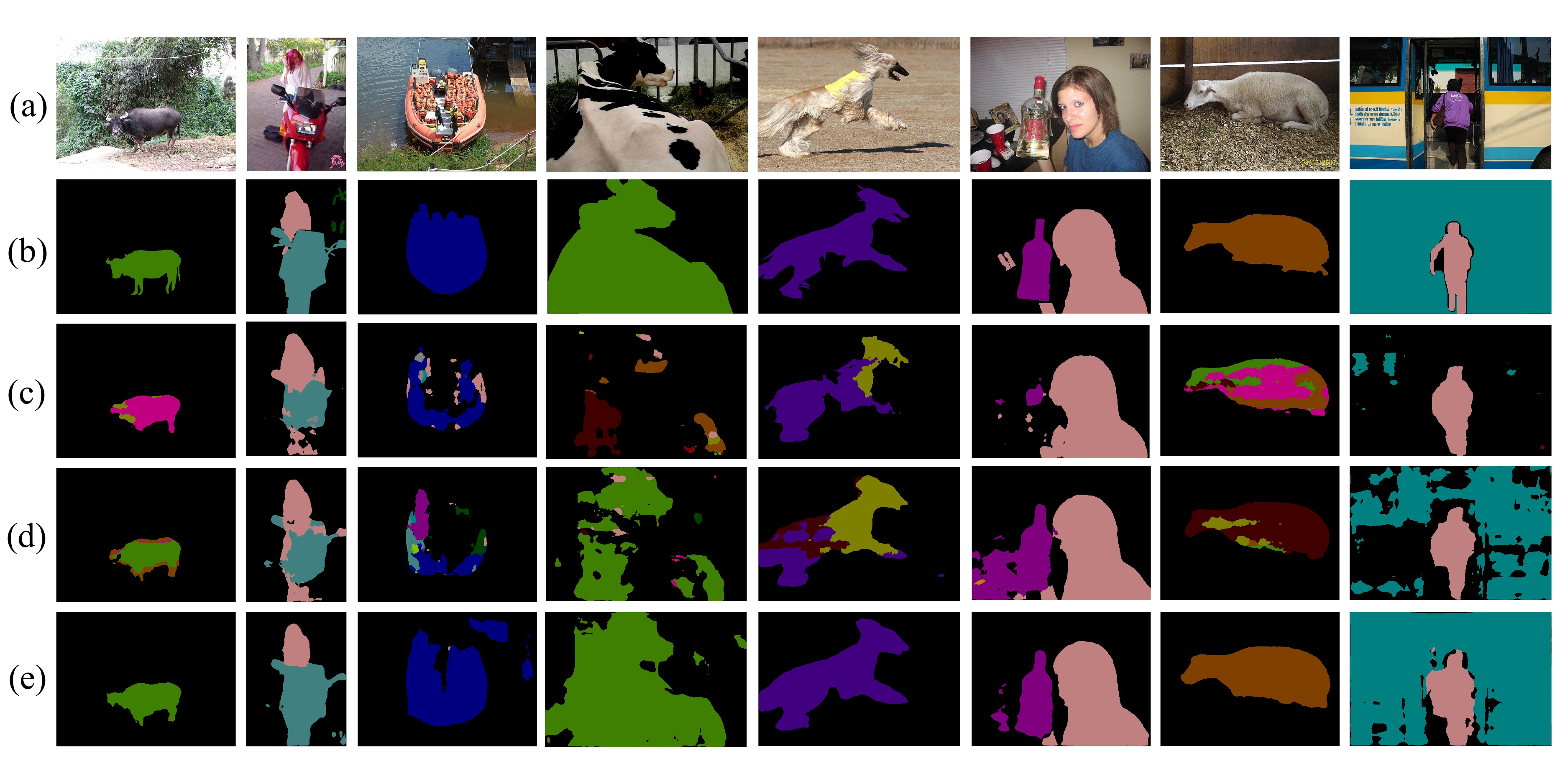}
\end{center}
\caption{Qualitative results on the PASCAL VOC2012 Val. (a) Input images. (b) Ground truth labels. (c) The segmentation results from Sup. $l$. (d) The segmentation results from CPS~\cite{chen2021semi}. (e) The segmentation results from MGD}
\label{fig: seg_comp}
\end{figure*}

\textbf{Qualitative Results.}
As illustrated in Fig.~\ref{fig: seg_comp}, we compare our method with the competitive approach CPS~\cite{chen2021semi} and Sup. ${l}$ in terms of segmentation results on PASCAL VOC2012 Val~\cite{everingham2010pascal}. 
From the figure, we find that Sup. ${l}$ and CPS fail to capture the internal pixels of the target or mistake the object categories. On the contrary, our approach manages to obtain the integral masks closer to the ground truth labels, which further proves the superiority of the MGD scheme and its multi-level setting. 

\begin{table}[t!]
  \centering
  \caption{Analysis of different teacher setting on VOC2012 and Cityscapes under 1/8 partition with student ResNet-18. $T_D$: ResNet-101, $T_W$: Wide ResNet-34.}
  \begin{threeparttable}
  \resizebox{0.6\textwidth}{!}{
      \begin{tabular}{cc|cc|cc}
      \toprule
        \multicolumn{2}{c|}{\textbf{First Teacher}}&\multicolumn{2}{c|}{\textbf{Second Teacher}}& \multirow{2}{*}{\textbf{VOC}}&\multirow{2}{*}{\textbf{Cityscapes}} \\ 
        \cline{1-4}
      $T_D$ & $T_W$ & $T_D$ &  $T_W$  & &  \\ 
      \hline
      $\surd$ &  & & & 64.85 & 69.63 \\
        & $\surd$ & & & 64.36 & 68.98 \\
        $\surd$ &  & $\surd$ & & 66.14 & 71.19\\
      & $\surd$ & & $\surd$ & 65.78 & 70.92\\
      $\surd$ &  & & $\surd$ &  \textbf{68.29} & \textbf{72.65}\\
      \bottomrule
      \end{tabular}}
      \end{threeparttable}
      \label{tab: res_4}
  \end{table} 
  
\subsection{Ablation Study}
\textbf{Analysis of Teacher Setting.}
To evaluate the significance of the complementary teacher scheme, we employ different teacher model combinations to optimize the lightweight model.
As shown in Tab.~\ref{tab: res_4}, employing two teacher models in the distillation procedure is better than a single model.
Furthermore, adopting different structural-based teachers can achieve more impressive performance than leveraging the same combination, which indicates that teacher models with different architectures help to provide complementary concepts and success in optimizing lightweight models from different dimensions.

\begin{table*}[t!]
  \centering
  \caption{The performance with different loss setting on the VOC2012 and Cityscapes under 1/8 partition protocol with student ResNet-18.}
  \begin{subtable}[t]{0.5\textwidth}
    \caption{Results of different levels distillation.}
    \resizebox{\linewidth}{!}{
      \begin{tabular}{lc|cc|cc|cc}
        \toprule
        \multicolumn{2}{c|}{\textbf{Pixel-level}}&\multicolumn{2}{c|}{\textbf{Image-level}}& \multicolumn{2}{c|}{\textbf{Region-level}}& \multirow{2}{*}{\textbf{VOC}}&\multirow{2}{*}{\textbf{City}} \\ 
        \cline{1-6}
      \footnotesize{\textbf{$\mathcal{L}_{Sup}$}} &
      \footnotesize{\textbf{$\mathcal{L}_{P}$}} & \footnotesize{\textbf{$\mathcal{L}_{I}^{T_D}$}} &
      \footnotesize{\textbf{$\mathcal{L}_{I}^{T_W}$}} &
      \footnotesize{\textbf{$\mathcal{L}_{R}^{T_D}$}} & 
      \footnotesize{\textbf{$\mathcal{L}_{R}^{T_W}$}} &  & \\ 
      \hline
      $\surd$ &  & & & & & 55.65 & 60.71\\
      $\surd$ & $\surd$ & & & & & 63.61 & 69.05\\
      $\surd$ & $\surd$ & $\surd$ & & & & 66.01 & 71.03\\
      $\surd$ & $\surd$ & & & & $\surd$ & 66.02 & 71.15\\
      $\surd$ & $\surd$ & $\surd$ & & & $\surd$ & \textbf{68.29} & \textbf{72.65}\\
      $\surd$ & $\surd$ & $\surd$ & $\surd$ &  $\surd$ & $\surd$ & 67.51 &  71.96\\
      \bottomrule
      \end{tabular}}
      
      \label{tab: res_5}
    \end{subtable}
    \hspace{10pt}
    \begin{subtable}[t]{0.45\textwidth}
        \centering
        \caption{Performance of labeled and unlabeled data distillation losses.}
        \resizebox{0.9\linewidth}{!}{
            \begin{tabular}{lc|ccc|cc}
            \toprule
            \multicolumn{2}{c|}{\textbf{$\mathcal{L}_{label}$}}&\multicolumn{3}{c|}{\textbf{$\mathcal{L}_{unlabel}$}}&  \multirow{2}{*}{\textbf{VOC}}&\multirow{2}{*}{\textbf{City}} \\ 
            \cline{1-5}
            \footnotesize{\textbf{$\mathcal{L}_{Sup}$}} &
            \footnotesize{\textbf{$\mathcal{L}_{P}^l$}} & \footnotesize{\textbf{$\mathcal{L}_{P}^u$}} & \footnotesize{\textbf{$\mathcal{L}_{I}^{T_D}$}} & 
            \footnotesize{\textbf{$\mathcal{L}_{R}^{T_W}$}} &  & \\ 
            \hline
            $\surd$ & $\surd$ & & & & 58.24 & 63.98\\
            $\surd$ & $\surd$ & $\surd$ &  $\surd$ & $\surd$ & \textbf{68.29} & \textbf{72.65}\\
            \bottomrule
            \end{tabular}}
            
            \label{tab: res_10}
        \end{subtable}
        \label{tab: res_loss}
\end{table*}

\textbf{Analysis of Loss Setting.}
To investigate the distillation loss setting, we conduct extensive experiments and analyze the components in the multi-level distillation loss and the cooperative distillation loss in Tab.~\ref{tab: res_loss}. 
As shown in Tab.~\ref{tab: res_5} , the variant with the $\mathcal{L}_{P}^{u,l}$ can obtain a notable improvement than the supervised baseline. 
$\mathcal{L}_{I}^{u}$ and $\mathcal{L}_{R}^{u}$ further boost the performance of the lightweight model to 66.01/71.03\% and 66.02/71.15\%, which proves the effectiveness of region-level content-aware loss and image-level semantic-sensitive loss. 
We employ $\mathcal{L}_{I}^{u}$ and $\mathcal{L}_{R}^{u}$ on different branches to further explore the optimal configuration for these two losses. The experimental results show that when applying $\mathcal{L}_{I}^{u}$ to the deeper teacher model and $\mathcal{L}_{R}^{u}$ to a wider model can achieve the best results, \ie, 68.29\%/72.65\%. 
It may be because the deeper teacher is beneficial to provide the corresponding global semantic characteristics while the wider model has the advantage in capturing local context information. The results in Tab.~\ref{tab: res_5} also reveal that the distillation losses from image-level, region-level and pixel-level are all essential for breaking through the bottleneck of the capacity of the lightweight model.
As summarized in Tab.~\ref{tab: res_10}, MGD can achieve 10.05\% and 8.67\% performance improvement via unlabeled data distillation, which demonstrates the effectiveness of the unlabeled distillation loss. And it further reveals the labeled-unlabeled data cooperative distillation is efficient for optimizing the lightweight SSSS model.

\textbf{Analysis of Size of Regional Content Vectors.}
The regional content vectors $\mathbf{V} \in \mathcal{R}^{C \times H_v \times  W_v}$ are generated by the adaptive average pooling operation. To determine the optimal size of $H_v \times  W_v$, we follow the dataset characteristic to set $H_v=W_v$ and conduct the experiments on PASCAL VOC2012. As shown in Tab.~\ref{tab: res_8}, the MGD can achieve the best result when $H_v \times W_v = 7 \times 7$. 

\begin{table}[t!]
    \centering
    \caption{Different size of $H_v$ and $W_v$ on VOC2012 under 1/8 partition protocol with student ResNet-18.}
    \begin{threeparttable}
        \begin{tabular}{cccccc}
        \toprule
        \textbf{$H_v \times W_v$} & \textbf{$3 \times 3$} & \textbf{$5 \times 5$} & \textbf{$7 \times 7$} & \textbf{$9 \times 9$} & \textbf{$11 \times 11$}   \\ 
        \midrule
        mIoU & 66.83 & 67.01 & \textbf{68.29} & 67.75 & 67.38 \\
        \midrule
        \end{tabular}
        \end{threeparttable}
        \label{tab: res_8}
    \end{table}


\section{Discussion}
In this paper, we provide an effective solution on how to obtain a lightweight segmentation model in a semi-supervised setting. 
Furthermore, there are still many aspects worthy of further study: i) How to formulate the whole optimization process into a one-stage procedure? ii) How to further narrow the performance gap between the lightweight model and the large model?
iii) Is there any data selection strategy such as the active learning algorithm that can improve the performance?
We hope that our work can provide inspired insight and encourage more researchers to involve in the field of lightweight semi-supervised models.

\section{Conclusion}
In this paper, we offer the first attempt to obtain lightweight SSSS models via a novel multi-granularity distillation (MGD) scheme.
MGD employs triplet-branches to distill diverse granularity concepts from two complementary teacher models into a student one. Within MGD, the labeled-unlabeled data cooperative distillation scheme is set up to take full advantage of semi-supervised data characteristics.
Furthermore, MGD designs a hierarchical and multi-level loss setting for unlabeled data, which consists of an image-level semantic-sensitive loss, a region-level content-aware loss, and a pixel-level consistency loss. 
This novel hierarchical loss setting plays a key role to break through the bottleneck of the capacity of the lightweight model.
We conduct extensive experiments on two benchmarks to demonstrate the effectiveness of the proposed approach and analyze the key factors that contribute more to addressing this task.

\clearpage
%
%
\bibliographystyle{splncs04}
\bibliography{MGD}
\end{document}